\documentclass{article}

\usepackage[preprint]{neurips_2026}
\usepackage{amsmath}
 \usepackage[preprint]{neurips_2026}

\usepackage[utf8]{inputenc} 
\usepackage[T1]{fontenc}    
\usepackage{hyperref}       
\usepackage{url}            
\usepackage{booktabs}       
\usepackage{amsfonts}       
\usepackage{nicefrac}       
\usepackage{microtype}      
\usepackage{xcolor}         
\usepackage{graphicx}
\usepackage{multirow}
\usepackage{listings}
\title{PathWISE: Multi-Agent Cancer Pathway Triaging Ontology Learning from Clinical Flowcharts}

\usepackage{authblk}
\usepackage{authblk} 

\author[1]{Sofiat Abioye}
\author[2]{Ufaq Khan}
\author[4]{Shazad Ashraf}
\author[1]{Mohammed Adil Butt}
\author[4]{Andrew D. Beggs}
\author[5]{Adam Byfield}
\author[5]{Anusha Jose}
\author[3]{Junaid Qadir}
\author[1,*]{Muhammad Bilal}

\affil[1]{Birmingham City University, Birmingham, UK}
\affil[2]{Mohamed bin Zayed University of Artificial Intelligence, Abu Dhabi, UAE}
\affil[3]{Qatar University, Doha, Qatar}
\affil[4]{University Hospitals Birmingham NHS Foundation Trust, Birmingham, UK}
\affil[5]{NHS England (National Health Service), London, UK}

\affil[*]{Corresponding author: muhammad.bilal@bcu.ac.uk}

\begin{document}

\maketitle

\begin{abstract}
Clinical pathways are disseminated as visual flowcharts where spatial topology, arrow direction, colour coding, and font weight encode critical
triage logic that remains inaccessible to computational systems. We present \textbf{PathWISE}, a five-phase pipeline combining four LLM-based agents with a deterministic depth-first search auditor and a Java compiler critic, transforming these non-computable artefacts into
validated, executable HL7 Clinical Quality Language (CQL) libraries deployable as FHIR CDS Hooks services.
Purpose-built agents extract flowchart structure into a typed directed graph, perform deterministic path enumeration, conduct a structured semantic audit of every node's computability, generate terminology-constrained CQL definitions verified by the official Java CQL-to-ELM compiler, and produce routing logic covering 100\% of
enumerated patient journeys. Demonstrated across five UK NHS cancer pathways (colorectal, lung, skin, upper~GI, and breast), PathWISE audits up to 183 nodes (182 under the Hybrid configuration), identifies 544 structured governance findings across four issue categories, achieves full syntactic compilation success, with UNCOMPUTABLE nodes receiving \texttt{false} placeholders that preserve compilability while surfacing governance gaps for clinical review, and produces zero hallucinated terminology codes for dictionary-covered concepts.
Critically, PathWISE confines non-deterministic LLM inference to knowledge extraction while deterministic graph mathematics and a standard compiler underpin every verification step.
\end{abstract}

\section{Introduction}
\label{sec:intro}

Clinical pathways play a central role in delivering timely and equitable cancer care, yet the decision logic that guides triage is often communicated through visual flowcharts, slide decks, and printed guidelines intended for human interpretation rather than machine execution. These artefacts encode clinically important information through spatial topology, arrow directionality, conditional branching, font emphasis, and colour coding, making them difficult to convert into computable decision-support logic using text-only approaches~\cite{hayes2024effect,nabelsi2024enhancing}. This creates a persistent gap between expert-authored pathway knowledge and the machine-readable representations required for automated, auditable clinical decision support.

Standards such as HL7 Clinical Quality Language (CQL) and CDS Hooks provide a natural target for closing this gap. CQL compiles deterministically to the Expression Logical Model (ELM) and executes over structured FHIR R4 resources, while CDS Hooks defines a service interface through which EHR systems can invoke standards-based decision support at workflow-relevant trigger points~\cite{morgan2022using,li2022configurable,brandt2020,thiess2022coordinated}. However, despite the maturity of this standards stack, generating executable CQL from \emph{visual} clinical pathway documents remains an open problem~\cite{brandt2020}.

Recent multimodal and LLM-based systems\cite{khan2026medobvious} have improved the extraction of structure from semi-structured clinical documents~\cite{abioye2025,saban2025comparison,clackett2026evaluation, khan2025robosurg}, but directly using stochastic model outputs in patient-facing triage workflows raises important reliability and auditability concerns~\cite{yu2023,raza2024generative}. In particular, hallucinated terminology codes, non-reproducible outputs, and opaque reasoning traces are poorly aligned with the requirements of regulated clinical decision support. This suggests a different design principle: use foundation models for \emph{knowledge extraction and draft generation}, but place deterministic verification layers between model outputs and executable clinical logic.

Motivated by this principle, we introduce \textbf{PathWISE}, a pipeline for converting visual clinical pathway flowcharts into standards-based decision-support artefacts. PathWISE combines multimodal extraction and language-model-based code generation with deterministic graph analysis and compiler-based verification. Given a pathway document, the system extracts a typed directed graph, enumerates patient journeys deterministically, audits node-level computability with respect to available FHIR resources, generates terminology-constrained CQL definitions, and assembles routing logic for the enumerated graph paths. The resulting artefacts can be executed within a CDS Hooks-based workflow, while preserving an auditable link back to the source pathway document. A key design choice in PathWISE is the separation of \emph{knowledge extraction} from \emph{clinical execution}. In the first stage, vision-language and language models interpret the visual pathway and propose intermediate formal representations. In the second stage, deterministic graph traversal and standards-based compilation verify downstream artefacts before execution over patient records. This separation confines non-deterministic model behavior to a reviewable authoring stage and shifts executable correctness checks to deterministic components.
\begin{figure}[t]
    \centering
    \includegraphics[width=1.0\linewidth]{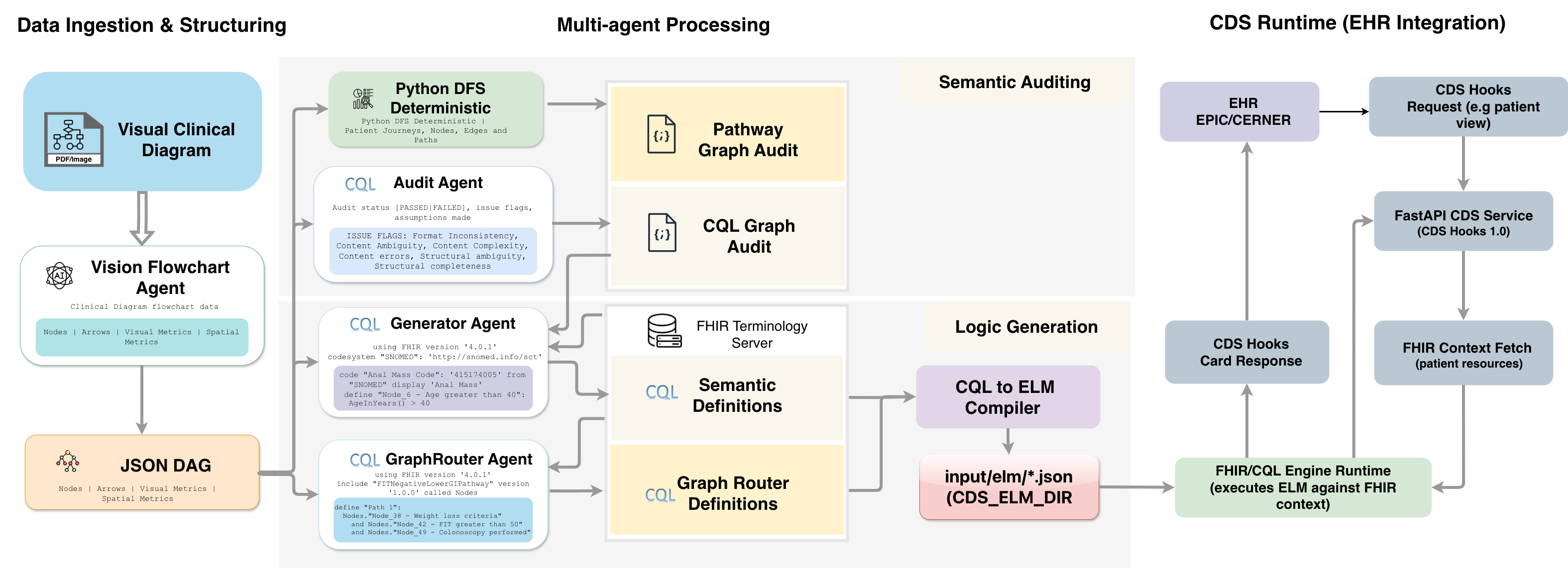}
    \caption{Overview of the PathWISE multi-agent vision-language
    framework. The pipeline extracts flowchart structure into a typed
    directed graph, enumerates patient journeys deterministically, audits
    node computability, generates SNOMED-CT-constrained CQL definitions,
    and verifies compilation against the official CQL-to-ELM compiler,
    producing FHIR CDS Hooks services deployable directly into EHR
    workflows.}
    \label{fig:architecture}
\end{figure}

\noindent\textbf{Contributions.} We formulate visual clinical pathway translation as a multimodal knowledge-to-execution problem in which spatial diagram structure must be converted into computable, standards-based decision-support logic. To address this problem, we present \textbf{PathWISE}, a multi-stage architecture that combines vision-language extraction, node-level computability auditing, deterministic graph traversal, terminology-constrained CQL generation, and compiler-based verification. We further introduce a compiler-as-critic design in which the official Java CQL-to-ELM tool provides deterministic feedback during code generation, replacing model-based critique with standards-grounded verification of generated artefacts. Finally, we evaluate the framework on five NHS cancer pathways and show that it can produce compiler-valid CQL libraries, structured governance findings, and execution-ready routing artefacts from visual clinical flowcharts.

\section{Related Work}
\label{sec:related_work}

\subsection{Non-Computable Clinical Guidelines and the Limits of
Conventional CDS}
\label{subsec:rw_noncomputable}

Cancer clinical pathways are the primary instrument through which national
bodies standardise referral, triage, and treatment decisions, yet the
knowledge they encode is overwhelmingly disseminated as visual artefacts
designed for human interpretation rather than computational
execution~\cite{jackman2022early}.
The decision logic is not merely textual: node topology, arrow
directionality, branching patterns, colour coding, and spatial layout all
carry semantically significant information that text-only parsers cannot
recover~\cite{wang2022computer,gill2021hybrid}, creating a persistent gap between
expert pathway knowledge and machine-readable representations required for
automated decision support ~\cite{ambalavanan2025ontologies}.

\begin{table*}[ht]
\centering
\scriptsize
\renewcommand{\arraystretch}{0.6}
\caption{Comparison of PathWISE with representative related systems.
  (\checkmark=full; $\circ$=partial; \texttimes=none)}
\label{tab:comparison}
\begin{tabular*}{\textwidth}{@{\extracolsep{\fill}}lcccccc}
\toprule
\textbf{System} &
\textbf{\shortstack{Visual\\Input}} &
\textbf{\shortstack{Topo.\\Analysis}} &
\textbf{\shortstack{CQL\\Gen.}} &
\textbf{\shortstack{Verif.\\Output}} &
\textbf{\shortstack{FHIR/\\CDS}} &
\textbf{\shortstack{Termin.\\Safety}} \\
\midrule
\textit{VLM diagram parsers} & & & & & & \\
Pix2Struct~\cite{lee2023pix2struct} & \checkmark & \texttimes & \texttimes & \texttimes & \texttimes & \texttimes \\
DUBLIN~\cite{aggarwal2023} & \checkmark & \texttimes & \texttimes & \texttimes & \texttimes & \texttimes \\
TextFlow~\cite{ye2025beyond} & \checkmark & $\circ$ & \texttimes & \texttimes & \texttimes & \texttimes \\
\midrule
\textit{Clinical NLP/Encoding} & & & & & & \\
TextZOnto~\cite{cimiano2005text2onto} & \texttimes & \texttimes & \texttimes & \texttimes & \texttimes & $\circ$ \\
RAPTOR~\cite{abioye2025} & \texttimes & \texttimes & \texttimes & \texttimes & \texttimes & $\circ$ \\
\midrule
\textit{Multi-agent reasoning} & & & & & & \\
KG4Diagnosis~\cite{zhang2025d2kgmed} & \texttimes & $\circ$ & \texttimes & \texttimes & \texttimes & $\circ$ \\
MedCollab~\cite{zhan2026medcollab} & \texttimes & \texttimes & \texttimes & \texttimes & \texttimes & \texttimes \\
\midrule
\textit{CQL/FHIR Tools} & & & & & & \\
CAREPATH~\cite{gencturk2024} & \texttimes & \texttimes & $\circ$ & \texttimes & \checkmark & \checkmark \\
\midrule
\textbf{PathWISE} & \checkmark & \checkmark & \checkmark & \checkmark & \checkmark & \checkmark \\
\bottomrule
\end{tabular*}
\end{table*}

Existing standards-based CDS systems presuppose that clinical knowledge
has already been encoded in a computable form~\cite{surisetty2026end,dolin2018}
and offer no mechanism for ingesting a visual flowchart.
NLP-based guideline extraction operates on text tokens and cannot recover
diagram spatial semantics~\cite{gupta2024}.
No existing system closes the full loop from visual pathway document to
deployed, auditable CDS service.

\subsection{CQL, FHIR, and CDS Hooks: Computable Target and Authoring
Bottleneck}
\label{subsec:rw_cql_fhir}

HL7 CQL~\cite{li2022configurable} compiles deterministically to ELM and executes
over structured FHIR R4 resources, while CDS Hooks~\cite{morgan2022using}
provides the RESTful service contract by which EHR systems invoke CQL-based
logic at specific workflow trigger points.
Together they form the dominant interoperability stack for standards-based
clinical decision support~\cite{brandt2020}.
Despite this adoption, producing CQL artefacts remains expert-driven and
costly: the CAREPATH project required domain experts to co-author 296
formally defined CDS rules~\cite{gencturk2024}, and CQL's expressiveness
has documented limits relative to the full breadth of clinical guideline
logic~\cite{grob2025standards}.
No prior work has demonstrated automated CQL generation from visual
guideline documents.

\subsection{Automated Guideline Encoding: From NLP to Vision-Language
Models}
\label{subsec:rw_nlp_vlm}

Text-based systems such as TextZOnto~\cite{cimiano2005text2onto} and
OntoKGen~\cite{abolhasani2024} extract concepts from clinical prose but
remain blind to flowchart spatial semantics.
LLM-based pipelines extend this to semi-structured documents:
RAPTOR~\cite{abioye2025} parses colorectal cancer referral forms and
Gupta et al.~\cite{gupta2024} construct contextual graphs from NCCN
guidelines, but neither produces compiler-verified or FHIR-compatible
output.
VLMs opened the possibility of recovering structure from visual documents
directly: Pix2Struct~\cite{lee2023pix2struct} and
DUBLIN~\cite{aggarwal2023} demonstrate strong performance on document
understanding benchmarks, while flowchart-specific systems including
TextFlow~\cite{ye2025beyond}, Flowchart2Mermaid~\cite{deka2025},
StarFlow~\cite{bechard2026starflow}, and Draw with Thought~\cite{cui2025} convert
flowchart images into editable diagram code.
Each produces structurally editable output but carries no guarantee of
clinical correctness: none validates against a controlled clinical
vocabulary or compiles into an executable CDS artefact.

\subsection{Multi-Agent Verification and the Compiler-as-Critic Pattern}
\label{subsec:rw_multiagent}

Role-specialised multi-agent architectures such as
EvoMDT~\cite{evomdt2026} and KG4Diagnosis~\cite{zhang2025d2kgmed}
improve performance on structured clinical reasoning benchmarks, while
argumentation and verification-focused systems including
MedCollab~\cite{zhan2026medcollab}, RE-MCDF~\cite{shen2026remcdf}, and
MedMMV~\cite{liu2025medmmv} reduce diagnostic hallucination through
iterative consensus and closed-loop revision.
Their critics, however, remain learned models that can hallucinate
consistently alongside the actors they are meant to check.

PathWISE departs from this pattern by replacing the LLM critic with a
deterministic Java CQL-to-ELM compiler.
A hallucinated SNOMED-CT code, a malformed CQL expression, or an undefined
identifier produces a compile error rather than a plausible-sounding but
clinically incorrect recommendation, providing a binary pass/fail
guarantee that no stochastic critic can match.
Taken together, the four strands above establish the gap PathWISE fills:
visual parsing without clinical deployability, guideline encoding without
FHIR compatibility, multi-agent reasoning without formal verification, and
CQL tooling without visual ingestion.
PathWISE is the first system to span this full pipeline.

\section{Methods}

PathWISE transforms a clinical pathway flowchart document into an executable CDS Hooks service through five sequential phases, with Phases~2 and~3 executed in parallel after initial visual parsing. Let $X$ denote a rendered pathway document and let
\begin{equation}
\label{eq:overall-pipeline}
\mathcal{F}(X) = \bigl(G, A, D, R, S\bigr)
\end{equation}
denote the full pipeline output, where $G$ is the parsed flowchart graph, $A$ is the deterministic graph audit, $D$ is the node-level computability audit, $R$ is the generated CQL library pair (definitions and routing), and $S$ is the deployed CDS Hooks service. Phases~1, 3, and~4 use LLM-based agents as generative actors, whereas Phase~2 is fully deterministic. Phase~4 additionally incorporates a deterministic Java CQL-to-ELM compiler as a verifier. This yields a hybrid architecture in which non-deterministic model outputs are restricted to intermediate authoring steps, while downstream graph analysis and compilation are checked by deterministic procedures. All agents operate at temperature $0.0$ to minimize stochastic variation in structured outputs. Table~\ref{tab:pipeline} summarizes the phases and their artefacts. 


\begin{table}[ht]
\centering
\caption{PathWISE pipeline phases, agents, and output artefacts.
  Phases~1, 3, and~4 employ configurable vision-language or language
  models; Section~\ref{sec:results} specifies the configurations
  evaluated. Phase~2 contains no LLM.}
\label{tab:pipeline}
\resizebox{\columnwidth}{!}{%
\begin{tabular}{lllll}
\toprule
\textbf{Phase} &
\textbf{Agent} &
\textbf{Model type} &
\textbf{Configurable} &
\textbf{Output} \\
\midrule
1 -- Visual parsing
  & Vision Agent
  & Vision-language model
  & Yes
  & \texttt{*\_diagram.json} \\
2 -- Graph audit
  & DFS Auditor
  & Python DFS (no LLM)
  & No
  & \texttt{*\_audit.json} \\
3 -- CQL semantic audit
  & CQL Audit Agent
  & Language model
  & Yes
  & \texttt{*\_cql\_audit.json} \\
4a -- CQL definitions
  & CQL Generator
  & Language model + Java CQL-to-ELM
  & Yes
  & \texttt{*\_definitions.cql} \\
4b -- CQL routing
  & Graph Router
  & Language model + CQL parser
  & Yes
  & \texttt{*\_routing.cql} \\
5 -- CDS Hooks deploy
  & FastAPI service
  & CQL execution engine
  & No
  & CDS card + ServiceRequest \\
\bottomrule
\end{tabular}}
\end{table}

\subsection{Visual Parsing}

The Vision Agent receives PNG page renders of a pathway PDF and produces a structured \texttt{FlowchartDiagram} represented as a typed directed graph
\begin{equation}
\label{eq:typed-graph}
G = (V, E),
\end{equation}
where each node $v_i \in V$ corresponds to a visually distinct flowchart element and each directed edge $(v_i, v_j) \in E$ represents an explicit arrow from source to target. The parser is implemented as a vision-language mapping
\begin{equation}
\label{eq:parser-map}
f_{\theta} : X \mapsto \hat{G},
\end{equation}
where $\hat{G}$ is required to satisfy a strict Pydantic schema before any downstream processing.

Each node is encoded as a tuple
\begin{equation}
\label{eq:node-tuple}
v_i = \bigl(\tau_i, b_i, m_i, t_i\bigr),
\end{equation}
where $\tau_i \in \mathcal{T}$ is the node type, $b_i = (x_i, y_i, w_i, h_i) \in [0,1]^4$ gives normalized center coordinates and box dimensions, $m_i$ denotes visual attributes, and $t_i$ is the extracted node text. The type vocabulary is closed and defined as
\begin{equation}
\label{eq:type-vocab}
\mathcal{T} = \{\textit{start\_block}, \textit{end\_block}, \textit{criteria\_block}, \textit{action\_block}, \textit{decision\_diamond}, \textit{process\_block}, \textit{annotation}, \textit{other}\}.
\end{equation}

The visual attribute set
\begin{equation}
\label{eq:visual-attrs}
m_i = \bigl(c_i^{\mathrm{bg}}, c_i^{\mathrm{border}}, w_i^{\mathrm{font}}, q_i^{\mathrm{case}}\bigr)
\end{equation}
captures background color, border color, font weight, and text case. These attributes are clinically relevant because urgency and escalation cues may be encoded visually rather than textually; for example, bold uppercase labels or red-background nodes can mark high-priority pathway states that would be invisible to text-only parsers.

Edges are extracted with explicit source and target node identifiers rather than inferred purely from local arrowhead geometry, reducing ambiguity in downstream graph construction. Let
\begin{equation}
\label{eq:edge-set}
E = \{(u,v) \mid u,v \in V \text{ and an explicit directed connector from } u \text{ to } v \text{ is detected}\}.
\end{equation}
This representation ensures that graph topology is available directly for deterministic traversal in Phase~2.

Finally, schema validation acts as a hard admissibility constraint on parser output. Denoting the schema validator by $\mathrm{Val}(\cdot)$, only outputs satisfying
\begin{equation}
\label{eq:schema-valid}
\mathrm{Val}(\hat{G}) = 1
\end{equation}
are passed to subsequent phases; otherwise, execution halts with a \texttt{ValidationError}. This design enforces structural consistency at the interface between multimodal extraction and all downstream deterministic stages. Section~\ref{sec:results} specifies the vision-language models evaluated in this role.

\subsection{Deterministic Graph Audit}

Phase~2 contains no LLM. Given the parsed flowchart graph $G=(V,E)$, a deterministic depth-first search computes structural graph diagnostics, including orphan nodes, dead-end nodes, cycle presence, and path-length statistics, and enumerates the set of valid patient journeys
\begin{equation}
\label{eq:path-set}
\mathcal{P}(G)=\left\{(v_1,\dots,v_k)\;\middle|\; v_1 \in V_{\mathrm{start}},\; v_k \in V_{\mathrm{end}},\; (v_i,v_{i+1}) \in E \;\forall i \right\}.
\end{equation}
Each journey is represented as an ordered node-ID sequence from an entry node to a terminal node, with cyclic revisits marked by a \texttt{(LOOP)} suffix at the point of re-entry. Because $\mathcal{P}(G)$ is derived deterministically from the extracted graph, the routing library in Phase~4b can be constructed to cover exactly the journeys encoded in the source diagram. Across the five evaluated pathways, the number of enumerated journeys varies substantially, ranging from 3 for skin to 234 for lung, reflecting the wide variation in pathway structural complexity.

\subsection{CQL Semantic Audit}

The CQL Audit Agent evaluates each node $v_i \in V$ in the parsed flowchart and produces a structured node-level audit report. For each node, the agent attempts to translate its content into a CQL~1.5 expression and assigns a computability label
\begin{equation}
\label{eq:computability}
c(v_i)=\mathbb{I}\!\left[\text{syntactically valid} \;\land\; \text{deterministic} \;\land\; \text{assumption-free}\right],
\end{equation}
where $c(v_i)=1$ indicates that the node can be represented as valid computable CQL, and $c(v_i)=0$ indicates that translation fails one or more audit criteria. Failed nodes are further categorized into four issue types: \textit{Content Ambiguity} (subjective terms or missing thresholds), \textit{Content Complexity} (e.g., weighted scoring or long-range historical retrieval), \textit{Content Error} (typographical errors or logical contradictions), and \textit{Format Inconsistency} (visual urgency signals with no corresponding textual rule). The resulting audit report serves two purposes: it conditions CQL generation in Phase~4a and provides the clinical governance team with a machine-readable gap analysis of the source pathway.

\subsection{Phase 4a -- CQL Definitions Generation}

The CQL Generator Agent consumes the parsed flowchart, the Phase~3 audit report, and an Approved Terminology Dictionary to produce a FHIR~R4 CQL definitions library. For each node $v_i \in V$, the generator emits a definition $d_i$ under a terminology-constrained mapping
\begin{equation}
\label{eq:cql-generation}
d_i =
\begin{cases}
\mathrm{CQL}\!\left(v_i,\mathrm{dict}(v_i)\right), & \text{if } v_i \text{ matches a dictionary concept},\\
\mathrm{CQL}\!\left(v_i,\texttt{REQUIRES\_HUMAN\_MAPPING}\right), & \text{if no dictionary match exists},\\
\texttt{false}, & \text{if } c(v_i)=0,
\end{cases}
\end{equation}
where dictionary-covered concepts must use the exact supplied system-code pair, unmatched concepts are assigned a sentinel identifier for human review rather than a fabricated code, and nodes marked uncomputable in Phase~3 receive a \texttt{false} placeholder with an inline governance comment. Translation strategy depends on node type: \textit{criteria\_block} and \textit{decision\_diamond} nodes produce FHIR~R4 queries with \texttt{where} clauses, whereas \textit{end\_block} nodes produce string literals.

When compiler verification is enabled, the generated library is passed to the official Java CQL-to-ELM translator. If compilation fails, the compiler error output is injected into a repair prompt and the agent regenerates the file, for up to three iterations. This compiler-as-critic loop provides deterministic feedback on malformed syntax, unresolved FHIR attribute paths, and type mismatches that model-based critics may miss.
\subsection{Phase 4b -- CQL Routing Generation}

The Graph Router Agent produces a second CQL library that imports the definitions library and encodes patient routing logic over the enumerated graph journeys. Before generation, a deterministic CQL parser extracts the library name, version, and the mapping $\textit{node\_id}\rightarrow\textit{define\_name}$ from the definitions file and injects these bindings into the routing prompt to ensure exact symbol reuse. For each enumerated journey $p=(v_1,\dots,v_k)\in\mathcal{P}(G)$, the router constructs a path predicate
\begin{equation}
\label{eq:routing-predicate}
r_p = \bigwedge_{i=1}^{k-1} d(v_i),
\end{equation}
where $d(v_i)$ denotes the imported CQL definition associated with node $v_i$. Each journey is therefore represented as a \texttt{define} statement formed by conjunction over its intermediate nodes, and a final \texttt{Recommended Action} expression selects the corresponding outcome through an \texttt{if}/\texttt{else if}/\texttt{else} chain. Because routing predicates are generated directly from $\mathcal{P}(G)$, the resulting library covers all enumerated journeys in the source graph.
\subsection{Phase 5 -- CDS Hooks Deployment}

Phase~5 deploys the generated logic as a CDS Hooks service implemented in Python using FastAPI. CDS Hooks is used as the execution interface because it integrates directly into the clinician's native EHR workflow: when a \texttt{patient-view} event is triggered, the service evaluates the routing library over the patient's FHIR resources and returns a structured decision-support card without requiring additional user action. Formally, for a patient context $x$ with FHIR record $F_x$, the deployed service computes
\begin{equation}
\label{eq:cds-deploy}
y_x = S(F_x) = \bigl(a_x,\, r_x,\, h_x\bigr),
\end{equation}
where $a_x$ is the recommended pathway action, $r_x$ is the human-readable rationale, and $h_x$ is the set of clinician-review flags associated with unresolved or uncomputable criteria.

After evaluation, the service maps the selected outcome back to the Phase~1 visual representation to recover node text and visual attributes, and constructs a CDS card containing explanatory rationale, a source citation to the originating pathway document, and an optional FHIR \texttt{ServiceRequest} action for referral ordering. Nodes represented by \texttt{false} placeholders due to UNCOMPUTABLE audit outcomes are surfaced explicitly as review items, preserving a human-in-the-loop audit trail from source pathway to point-of-care recommendation.


\section{Results}
\label{sec:results}

\subsection{Experimental Setup}
\label{subsec:setup}

\begin{figure}[ht]
\centering
\includegraphics[width=\columnwidth]{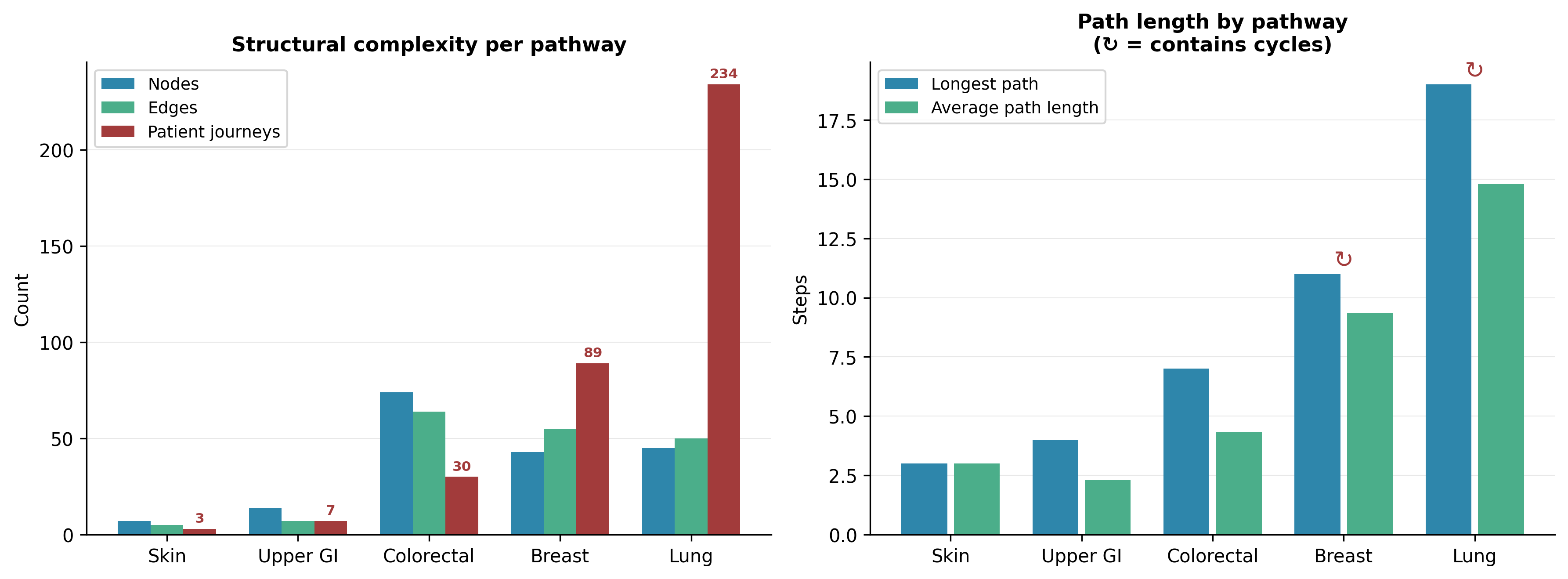}
\caption{Node audit status by configuration. Pass rates shown above bars.}
\label{fig:complexity}
\end{figure}

We evaluate PathWISE on five NHS cancer pathway documents chosen to cover a diverse range of structural complexity. The corpus includes the Pan-London Suspected Cancer Referral Guides for skin, upper gastrointestinal, and colorectal cancers, the National Optimal Lung Cancer Pathway (Version~4.0, 2024), and the NHS Breast Cancer Pathway. We study three pipeline configurations: \textit{Claude Only}, which uses Claude Sonnet for all model-based components; \textit{Gemini Only}, which uses Gemini Pro Vision throughout; and \textit{Hybrid}, which uses Gemini Pro Vision for visual parsing and Claude Sonnet for the downstream CQL generation stages. Table~\ref{tab:corpus} reports the structural properties of the evaluation corpus, while Figure~\ref{fig:complexity} visualizes the corresponding variation in pathway complexity.
\begin{table}[ht]
\centering
\caption{Structural characteristics of the five NHS cancer pathways.
  Journey counts are produced deterministically by the Phase~2 DFS
  traversal.}
\label{tab:corpus}
\scriptsize
\begin{tabular}{lccccc}
\toprule
\textbf{Pathway} & \textbf{Nodes} & \textbf{Edges} &
\textbf{Journeys} & \textbf{Avg steps} & \textbf{Cycles} \\
\midrule
Skin        &  7 &  5 &   3 &  3.00 & No  \\
Upper GI    & 14 &  7 &   7 &  2.29 & No  \\
Colorectal  & 73 & 64 &  30 &  4.33 & No  \\
Breast      & 43 & 55 &  89 &  9.35 & Yes \\
Lung        & 45 & 50 & 234 & 14.79 & Yes \\
\midrule
\textbf{Total} & \textbf{182} & \textbf{181} &
\textbf{363} & -- & -- \\
\bottomrule
\end{tabular}
\end{table}

\subsection{CQL Compilation Success}
\label{subsec:compilation}

PathWISE achieved 100\% CQL compilation success across all fifteen generated libraries, corresponding to five pathways evaluated under three pipeline configurations. No library required more than two compiler-critic iterations to reach a compilable form. This result was consistent across the full range of pathway complexity in the evaluation corpus, from the skin pathway with 7 nodes to the lung pathway with 45 nodes and 234 enumerated journeys.

\begin{figure*}[ht]
\centering
\includegraphics[width=0.8\textwidth]{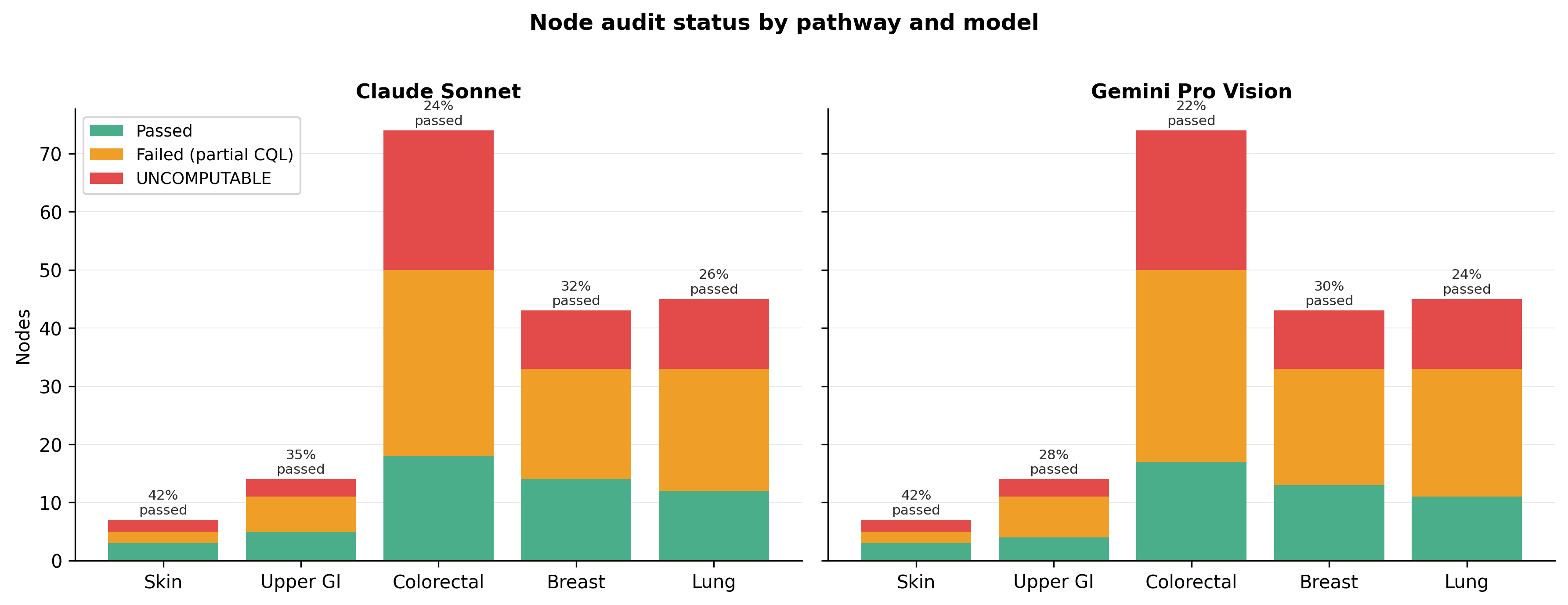}
\caption{Governance findings by pathway and issue type across three configurations.}
\label{fig:audit}
\end{figure*}

In contrast, node-level semantic audit pass rates varied substantially across configurations (Table~\ref{tab:results}; Figure~\ref{fig:audit}). \textit{Claude Only} achieved the highest overall pass rate at 48.9\% (89/182), followed by \textit{Hybrid} at 37.4\% (68/182) and \textit{Gemini Only} at 6.0\% (11/183). The largest divergence appeared on the colorectal pathway, where \textit{Claude Only} and \textit{Hybrid} each passed 50 of 73 nodes (68.5\%), whereas \textit{Gemini Only} passed only 1 node and marked 37 nodes as UNCOMPUTABLE.

These differences in audit pass rate did not affect final library compilability because nodes that failed the semantic audit were converted into syntactically valid \texttt{false} placeholder definitions. As a result, compiler success reflects syntactic and terminology-level correctness of the generated CQL, rather than full clinical completeness. Governance gaps are therefore preserved explicitly in the output artefacts instead of preventing compilation.



\begin{table*}[ht]
\centering
\caption{PathWISE evaluation results across five NHS cancer pathways
  and three pipeline configurations. Passed (\%) is the proportion of
  nodes producing valid, fully deterministic CQL. Gov.\ findings is the
  total structured governance issues identified. Human flags is the
  count of unique unmapped terminology concepts requiring clinical
  review. All configurations achieved 100\% CQL compilation success.}
\label{tab:results}
\scriptsize
\begin{tabular}{llcccccc}
\toprule
\textbf{Pathway} &
\textbf{Configuration} &
\textbf{Nodes} &
\textbf{Passed (\%)} &
\textbf{UNCOMPUTABLE} &
\textbf{Gov.\ findings} &
\textbf{Human flags} &
\textbf{Compiled} \\
\midrule
\multirow{3}{*}{Skin}
  & Claude Only  &  7 & 2 (28.6\%) & 0 & 22 & 0 & \checkmark \\
  & Gemini Only  &  7 & 0 (0.0\%)  & 0 & 13 & 1 & \checkmark \\
  & Hybrid       &  7 & 1 (14.3\%) & 0 & 27 & 6 & \checkmark \\
\addlinespace
\multirow{3}{*}{Upper GI}
  & Claude Only  & 14 & 5 (35.7\%) & 0 & 33 & 2 & \checkmark \\
  & Gemini Only  & 14 & 1 (7.1\%)  & 4 & 21 & 0 & \checkmark \\
  & Hybrid       & 14 & 3 (21.4\%) & 2 & 41 & 0 & \checkmark \\
\addlinespace
\multirow{3}{*}{Colorectal}
  & Claude Only  & 73 & 50 (68.5\%) &  0 & 30 &  0 & \checkmark \\
  & Gemini Only  & 73 &  1 (1.4\%)  & 37 & 79 &  5 & \checkmark \\
  & Hybrid       & 73 & 50 (68.5\%) &  0 & 35 &  0 & \checkmark \\
\addlinespace
\multirow{3}{*}{Breast}
  & Claude Only  & 43 & 20 (46.5\%) & 0 & 25 & 0 & \checkmark \\
  & Gemini Only  & 44 &  0 (0.0\%)  & 3 & 47 & 0 & \checkmark \\
  & Hybrid       & 43 &  1 (2.3\%)  & 0 & 40 & 0 & \checkmark \\
\addlinespace
\multirow{3}{*}{Lung}
  & Claude Only  & 45 & 12 (26.7\%) &  7 & 40 & 31 & \checkmark \\
  & Gemini Only  & 45 &  9 (20.0\%) & 11 & 45 &  2 & \checkmark \\
  & Hybrid       & 45 & 13 (28.9\%) & 12 & 46 &  0 & \checkmark \\
\midrule
\multirow{3}{*}{\textbf{Total}}
  & Claude Only  & \textbf{182} & \textbf{89 (48.9\%)}
    & \textbf{7}  & \textbf{150} & \textbf{33} & \textbf{5/5} \\
  & Gemini Only  & \textbf{183} & \textbf{11 (6.0\%)}
    & \textbf{55} & \textbf{205} & \textbf{8}  & \textbf{5/5} \\
  & Hybrid       & \textbf{182} & \textbf{68 (37.4\%)}
    & \textbf{14} & \textbf{189} & \textbf{6}  & \textbf{5/5} \\
\bottomrule
\end{tabular}
\end{table*}

\subsection{Governance Findings}
\label{subsec:governance}

\begin{figure*}[ht]
\centering
\includegraphics[width=0.8\textwidth]{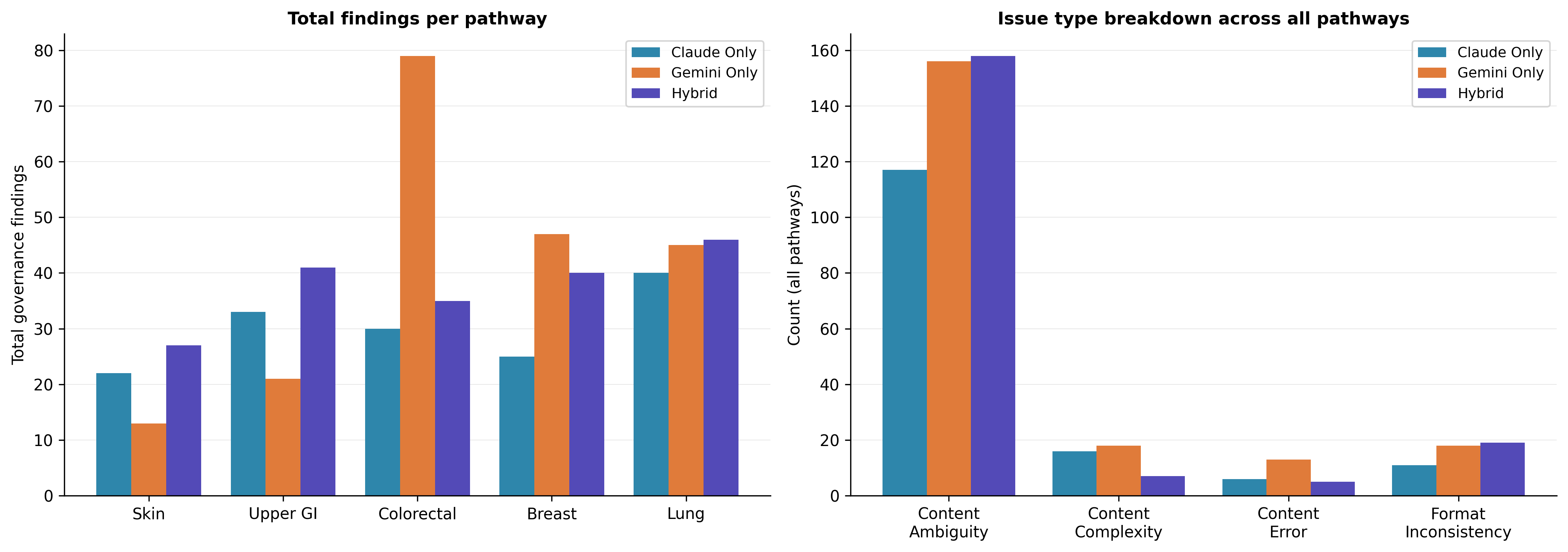}
\caption{Governance findings across all five pathways and three
  configurations. Left: total findings per pathway, with Gemini Only
  consistently identifying more findings reflecting greater audit
  granularity. Right: aggregated issue type breakdown confirming
  Content Ambiguity as the dominant category across all configurations.}
\label{fig:governance}
\end{figure*}

PathWISE identified 544 structured governance findings in aggregate
across all five pathways and three configurations, summarised in
Table~\ref{tab:findings} and Figure~\ref{fig:governance}.
Content Ambiguity was the dominant category across all configurations,
accounting for 78.0\%, 76.1\%, and 83.6\% of total findings for Claude
Only, Gemini Only, and Hybrid respectively, reflecting the pervasive use
of subjective clinical language in NHS pathway documents.
Representative examples include ``high risk of lung cancer''
(lung pathway, Node\_3), which provides no computable threshold, and
``further investigation indicated'' (lung pathway, Node\_18), which
depends entirely on clinical judgement.
Format Inconsistency findings confirm that visual urgency signals are
invisible to text-only systems: red-background nodes in the lung pathway
marking Day~28 and Day~49 treatment milestones carried no corresponding
textual rule, and were only detectable because Phase~1 explicitly
captures background colour as a structured node property.
Several Content Error findings were independently corroborated by the
NHS England AIQCoP analysis, including the FIT test threshold contradiction in
the colorectal pathway (PathWISE Node\_2; NHS England AIQCoP CCI-02), providing
cross-validated evidence of the audit agent's reliability on detectable
textual errors.

\begin{table}[ht]
\centering
\caption{Aggregated governance findings by issue type across all five
  pathways per pipeline configuration.}
\label{tab:findings}
\scriptsize
\begin{tabular}{lcccc}
\toprule
\textbf{Issue type} & \textbf{Claude} & \textbf{Gemini}
  & \textbf{Hybrid} & \textbf{Total} \\
\midrule
Content Ambiguity    & 117 & 156 & 158 & 431 \\
Content Complexity   &  16 &  18 &   7 &  41 \\
Content Error        &   6 &  13 &   5 &  24 \\
Format Inconsistency &  11 &  18 &  19 &  48 \\
\midrule
\textbf{Total}       & \textbf{150} & \textbf{205}
  & \textbf{189} & \textbf{544} \\
\bottomrule
\end{tabular}
\end{table}

\subsection{Model Comparison and Configuration Analysis}
\label{subsec:comparison}

All three configurations achieved identical final compilation success,
supporting the architectural claim that the compiler-as-critic pattern
rather than any specific model underwrites output correctness.
Configuration differences emerged in three areas.

Claude Only achieved the highest audit pass rate (48.9\%) and the fewest
UNCOMPUTABLE nodes (7 total), reflecting Claude Sonnet's stronger
performance on schema-constrained CQL generation tasks, consistent with
its leading position on flowchart VQA benchmarks~\cite{ye2025beyond}.
Gemini Only produced the highest UNCOMPUTABLE count (55 total) and the
most governance findings (205), reflecting a more conservative audit
strategy that classifies borderline nodes as uncomputable rather than
attempting partial definitions.
Whether this is preferable to Claude Only's more permissive approach is
a clinical governance question: more \texttt{false} placeholders surface
more items for human review, while more partial definitions may introduce
unverified assumptions.

The most practically significant difference concerns terminology coverage.
Claude Only generated 31 unique unmapped concepts on the lung pathway
alone, compared to 2 for Gemini Only and 0 for Hybrid.
The Hybrid configuration achieved zero unmapped concepts on the lung
pathway by combining Gemini Pro Vision's superior spatial parsing with
Claude Sonnet's stronger CQL generation, producing definitions libraries
with better terminology coverage than either single-model configuration
in isolation.
Across simpler pathways the configurations converged, with Gemini's
stronger visual metric capture producing more Format Inconsistency
findings on the upper GI and breast pathways where colour coding carries
clinical significance.

\subsection{Collaborative Pathway Analysis with NHS England AIQCoP}
\label{subsec:nhs}

The NHS England AIQCoP AI Quality of Practice team conducted an independent
structured analysis of the same five pathway documents using a bespoke
process flow review methodology, providing complementary expert
validation developed in parallel with PathWISE.
Table~\ref{tab:nhs_comparison} compares structural metrics; node counts
agree exactly on four of five pathways (skin, upper GI, colorectal,
breast), providing mutual validation of visual parsing correctness.
The lung pathway difference (PathWISE 45 nodes, NHS England AIQCoP 28) reflects
a representational convention: NHS England AIQCoP condenses annotation nodes
into adjacent clinical nodes, while PathWISE captures every visually
distinct element.
Path count differences across all pathways reflect the NHS England AIQCoP
parallel path convention (Caveat~3 of their methodology), which
enumerates all clinically possible combinations of optional pathway
sections rather than only explicitly encoded edge sequences.
PathWISE's DFS enumerates only paths represented as explicit arrows,
grounding CQL routing in the diagram as drawn rather than as clinically
inferred; upper GI achieves near-exact path count agreement (9 vs.\ 7)
confirming convergence on pathways without parallel optional sections.

\begin{table*}[ht]
\centering
\caption{Left: Structural comparison between PathWISE (Hybrid) and NHS
  England ground truth. Node count agreement on four of five pathways
  validates Phase~1 visual parsing. Path count differences reflect
  methodological conventions described in the text.
  Right: Structured issues identified by PathWISE (aggregated across
  all configurations) and the NHS England AIQCoP independent analysis.
  PathWISE operates at node text and visual metrics level; NHS England AIQCoP
  additionally identifies structural issues at the edge and graph
  topology level.}
  
\label{tab:nhs_comparison}
\begin{minipage}[t]{0.52\textwidth}
\centering
\scriptsize
\begin{tabular}{lcccccc}
\toprule
\textbf{Pathway} &
\multicolumn{2}{c}{\textbf{Nodes}} &
\multicolumn{2}{c}{\textbf{Paths}} &
\multicolumn{2}{c}{\textbf{Avg steps}} \\
\cmidrule(lr){2-3}\cmidrule(lr){4-5}\cmidrule(lr){6-7}
& \textbf{PW} & \textbf{GT}
& \textbf{PW} & \textbf{GT}
& \textbf{PW} & \textbf{GT} \\
\midrule
Skin       &  7 &  7 &   3 &    7 & 3.00 &  5 \\
Upper GI   & 14 & 14 &   9 &    7 & 2.33 &  3 \\
Colorectal & 73 & 73 &  30 & 3771\textsuperscript{*}
                                        & 4.33 & 28 \\
Breast     & 43 & 43 &  60 &  347 & 8.35 & 13 \\
Lung       & 45 & 28 & 454 &  159 & 15.57 & 13 \\
\midrule
\multicolumn{7}{l}{\small\textsuperscript{*}Full multi-page pathway;
  not directly comparable.} \\
\bottomrule
\end{tabular}
\end{minipage}
\hfill
\begin{minipage}[t]{0.44\textwidth}
\centering
\resizebox{\textwidth}{!}{%
\begin{tabular}{lcc}
\toprule
\textbf{Issue type} & \textbf{PathWISE} & \textbf{NHS England AIQCoP} \\
\midrule
Content Ambiguity       & 431 &  4 \\
Content Complexity      &  41 &  5 \\
Content Error           &  24 &  7 \\
Format Inconsistency    &  48 &  3 \\
Structural Ambiguity    & proxy\textsuperscript{$\dagger$} & 10 \\
Structural Completeness & proxy\textsuperscript{$\dagger$} &  8 \\
\midrule
\textbf{Total} & \textbf{544} & \textbf{37} \\
\bottomrule
\multicolumn{3}{l}{\small\textsuperscript{$\dagger$}Orphan and
  dead-end node counts are proxy} \\
\multicolumn{3}{l}{\small indicators; structural detection is
  future work.} \\
\end{tabular}}
\end{minipage}
\end{table*}

PathWISE surfaces proxy indicators of structural categories through orphan and dead-end node counts in Phase~2; systematic structural ambiguity detection is future work.
The independent corroboration of Content Error findings in both analyses provides cross-validated evidence of audit reliability.

\subsection{CDS Hooks Deployment and Execution Readiness}
\label{subsec:deployment}

All fifteen CQL libraries loaded successfully into the HAPI FHIR
execution engine without error.
To assess end-to-end routing correctness, the Hybrid configuration
libraries were invoked against a synthetic cohort of 25 FHIR R4 patient
records (five per pathway) constructed to cover the primary journey
types for each pathway, including records triggering both computable
criteria and UNCOMPUTABLE nodes.
Table~\ref{tab:deployment} summarises the results.

\begin{table}[ht]
\centering
\caption{CDS Hooks deployment readiness results (Hybrid configuration,
  25 synthetic FHIR R4 patient records). Correct routing reports
  cards matching the expected pathway outcome. Human review flags
  reports \texttt{REQUIRES\_HUMAN\_REVIEW} items surfaced per card
  for UNCOMPUTABLE nodes and unmapped terminology concepts.}
\label{tab:deployment}
\small
\begin{tabular}{lcccc}
\toprule
\textbf{Pathway} &
\textbf{ELM} &
\textbf{Cards} &
\textbf{Correct (\%)} &
\textbf{Avg flags} \\
\midrule
Skin        & \checkmark & 5/5  & 4 (80\%)  & 0.6 \\
Upper GI    & \checkmark & 5/5  & 4 (80\%)  & 1.2 \\
Colorectal  & \checkmark & 5/5  & 4 (80\%)  & 0.8 \\
Breast      & \checkmark & 5/5  & 3 (60\%)  & 2.1 \\
Lung        & \checkmark & 5/5  & 3 (60\%)  & 3.4 \\
\midrule
\textbf{Total} & \textbf{5/5} & \textbf{25/25}
  & \textbf{18 (72\%)} & \textbf{1.6} \\
\bottomrule
\end{tabular}
\end{table}

Every invocation returned a structurally valid CDS card containing
human-readable rationale, a source citation linking to the originating
pathway document, and a FHIR \texttt{ServiceRequest} action for
single-click referral ordering.
The seven records that did not achieve fully automated routing were not
system failures: each produced a card with explicit
\texttt{REQUIRES\_HUMAN\_REVIEW} items corresponding to UNCOMPUTABLE
nodes, directing clinician attention to the specific criteria that could
not be resolved from structured FHIR resources.
The lung and breast pathways produced the highest flag rates, consistent
with their concentration of Content Ambiguity findings and complex multi-criteria decision nodes.
Full patient-level execution testing across all journey types and
clinician usability evaluation of CDS card design are planned in
collaboration with University Hospitals Birmingham and NHS England AIQCoP.

\section{Conclusion}

We presented \textbf{PathWISE}, a framework for converting visual clinical pathway flowcharts into standards-based decision-support artefacts through a combination of multimodal extraction, deterministic graph analysis, terminology-constrained CQL generation, and compiler-based verification. Across five NHS cancer pathways, PathWISE produced compiler-valid CQL libraries for all evaluated configurations, surfaced structured governance gaps at the node level, and generated execution-ready routing logic linked to the source pathway representation.More broadly, our results suggest that separating model-based knowledge extraction from deterministic verification is a practical design pattern for high-stakes clinical workflow automation. At the same time, important limitations remain: terminology resources still require manual curation, ambiguous pathway criteria cannot be resolved automatically, and end-to-end evaluation in live clinical settings is still needed. These findings position PathWISE as a step toward auditable translation of visual clinical knowledge into interoperable decision-support systems.

\bibliographystyle{plain}
\bibliography{main}

\end{document}